# A Risk-aware Spatial-temporal Trajectory Planning Framework for Autonomous Vehicles Using QP-MPC and Dynamic Hazard Fields

Zhen Tian[1t], Zhihao Lin[1t], Dezong Zhao[1], *Senior Member, IEEE*, Christos Anagnostopoulos[2], *Member, IEEE*, Qiyuan Wang[2], Wenjing Zhao[1], Xiaodan Wang[3] and Chongfeng Wei[1] *Senior Member, IEEE*

*Abstract*—Trajectory planning is a critical component in ensuring the safety, stability, and efficiency of autonomous vehicles. While existing trajectory planning methods have achieved progress, they often suffer from high computational costs, unstable performance in dynamic environments, and limited validation across diverse scenarios. To overcome these challenges, we propose an enhanced QP-MPC-based framework that incorporates three key innovations: (i) a novel cost function designed with a dynamic hazard field, which explicitly balances safety, efficiency, and comfort; (ii) seamless integration of this cost function into the QP-MPC formulation, enabling direct optimization of desired driving behaviors; and (iii) extensive validation of the proposed framework across complex tasks. The spatial safe planning is guided by a dynamic hazard field (DHF) for risk assessment, while temporal safe planning is based on a space-time graph. Besides, the quintic polynomial sampling and sub-reward of comforts are used to ensure comforts during lane-changing. The sub-reward of efficiency is used to maintain driving efficiency. Finally, the proposed DHF-enhanced objective function integrates multiple objectives, providing a proper optimization tasks for QP-MPC. Extensive simulations demonstrate that the proposed framework outperforms benchmark optimization methods in terms of efficiency, stability, and comfort across a variety of scenarios likes lane-changing, overtaking, and crossing intersections.

## I. INTRODUCTION

IN recent years, Safe and stable autonomous driving relies heavily on effective trajectory planning. [1]. Throughout the driving process, trajectory planning enables the formulation of smooth, safe, and executable paths [2] based on the precise perception stages [3]–[5]. However, trajectory planning must simultaneously balance multiple factors [6],

*This work was supported in part by the EPSRC Innovation Fellowship of the Engineering and Physical Sciences Research Council of U.K. under Grant EP/S001956/2, in part by the Royal Society-Newton Advanced Fellowship under Grant NAF/R1/201213.

[1]Zhen Tian, Zhihao Lin, D. Zhao, W. Zhao, and C. Wei are with the School of Engineering, University of Glasgow, Glasgow, G12 8QQ, U.K. (e-mail: 2620920z@student.gla.ac.uk, 28004001@student.gla.ac.uk, dezong.zhao@glasgow.ac.uk, wenjing.zhao@glasgow.ac.uk, chongfeng.wei@glasgow.ac.uk).

[2]Christos Anagnostopoulos and Qiyuan Wang are with the School of Computing Science, University of Glasgow, Glasgow, G12 8RZ, U.K. (e-mail: Christos.Anagnostopoulos@glasgow.ac.uk, qiyuan.wang@glasgow.ac.uk).

[3]Xiaodan Wang is with the School of Engineering, Cardiff University, Cardiff, CF24 3AA, U.K. (e-mail: WangX223@cardiff.ac.uk).

t Equal contribution

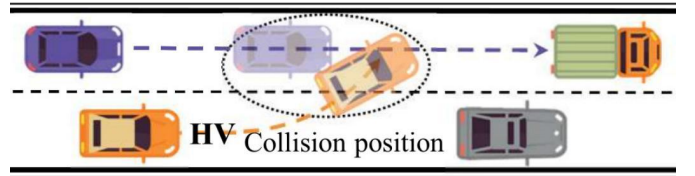

Fig. 1: How to make the autonomous vehicle interact with the vehicles beside it to make correct decisions and plan a reasonable trajectory to avoid collisions?

[7]. Achieving this balance driving remains challenging due to the broad decision-making scope in interactive scenarios. Some works directly use rule-based methods, contributing to absolute safety-based or efficiency-based decisions [8]. Consequently, efficiently planning safe, comfortable, and executable trajectories in intricate environments remains a significant challenge. In Fig. 1, the interaction between the host vehicle (HV) and surrounding vehicles is illustrated, highlighting the challenge of making correct decisions to avoid collisions. The HV must navigate complex interactive scenario, as shown by the potential collision position.

### A. Classical Planning Methods in Interactive Driving

To balance the driving considerations, a variety of trajectory planning methods have been proposed, including graph search methods, random sampling techniques, artificial potential fields, curve interpolation, and numerical optimization approaches [9]–[12]. Each of these methods offers unique advantages while also facing distinct limitations, shaping the ongoing evolution of trajectory planning research.

Graph search methods operate by constructing a representation of the environment using a graph, where nodes represent possible vehicle states and edges represent feasible transitions between states. The optimal trajectory is identified by minimizing a predefined cost function, typically based on criteria such as distance, time, or energy consumption [13]. One of the primary advantages of graph search methods is their ability to guarantee finding the optimal path under certain conditions. This advantage makes these methods well-suited for environments with discrete and well-defined state spaces. However, these methods often suffer from overly complex environment modeling, especially in dynamic or high-dimensional spaces, and exhibit low computational efficiency, limiting their practicality in real-time applications.

Random sampling methods, such as Rapidly-exploring Random Trees (RRT) and Probabilistic Roadmaps (PRM), generate random samples within the world space to explore feasible trajectories [14], [15]. These methods incrementally build a roadmap that satisfy safety and efficiency. Random sampling methods [16] offer probabilistic completeness, which means that if a solution exists, these methods have a high likelihood of finding it. Additionally, they are well-suited for tackling complex environments. Nevertheless, the computational efficiency of random sampling methods is still suboptimal, particularly in densely populated or constrained environments. Besides, the inherent randomness of the methods can lead to inconsistent planning and longer planning times.

Artificial potential field (APF) methods plan motion by simulating attractive forces towards the goal and repulsive forces away from obstacles [17]. By computing the resultant force at each step, the vehicle steers towards the goal while avoiding collisions. The APF method is mathematically straightforward and easy to implement, allowing for real-time obstacle avoidance without extensive computation. However, it is prone to getting stuck in local minima. In such a situation, the vehicle would be trapped in suboptimal positions while struggling with navigating around dynamic or multiple obstacles, potentially compromising trajectory quality.

Due to the limitations of the aforementioned methods, curve interpolation and numerical optimization techniques have emerged as solutions for trajectory planning. Curve interpolation methods determine the correlation coefficients of curves based on the vehicle's desired states. These methods generate trajectories that are excellent smoothness and feasibility. Bezier curves [18], B-splines [19], and quintic polynomials, have been commonly employed to ensure smooth transitions and adherence to kinematic constraints. On the other hand, numerical optimization approaches treat trajectory planning as an optimization problem minimizing objectives related to safety, smoothness, and efficiency while adhering to vehicle dynamics and collision avoidance constraints. Optimization frameworks such as Quadratic Programming (QP) and Nonlinear Programming (NLP) are utilized to solve for the optimal trajectory parameters [20], [21]. These methods are capable of producing high-quality, smooth, and dynamically feasible trajectories. However, these methods often involve high computational complexity. Additionally, accurate modeling of vehicle dynamics and environmental factors is essential, which can be challenging in fast-changing real-world situation.

Moreover, to enhance computational efficiency and reduce problem dimensionality, many studies decompose the three-dimensional trajectory planning problem into two two-dimensional planning tasks [22], [23]. These works typically involve decoupling path and speed planning within the Frenet coordinate system, thereby compressing the solution space and achieving faster computation time.

## B. Limitations of Existing Approaches

Traditional planning typically encounters challenges related to excessive computation, inconsistent planning, and difficulties in responding to dynamic obstacles and complex scenarios. Decoupling methods address excessive computation and poor responding to dynamic obstacles by separating path planning from speed planning. However, decoupling methods fail to fully exploit available computational resources, leading to potential suboptimality in complex scenarios. Additionally, decoupling methods often neglect the integration of speed curves during path planning, resulting in incompatible or suboptimal path configurations [22], [23]. Furthermore, current planning methods have been validated simple scenarios involving a few interacting vehicles, but their comparison with other benchmark optimization algorithms is still lacking [22], [24]. In light of these challenges, this study introduces a comprehensive framework for trajectory planning. The proposed framework integrates QP-Based Model Predictive Control (QP-MPC) with both spatial trajectory planning based on dynamic hazard field (DHF) risk assessment and temporal planning using a space-time graph. Besides, a novel objective function is proposed to include multiple assessments for the optimization, ensuring the safety, efficiency, and comforts. The proposed objective function also integrates the DHF, enabling the QP-MPC to account for risk distribution rather than merely providing a reference trajectory to the optimizer, as seen in existing optimization-based methods [22], [25]. By incorporating risk awareness,into optimization module, the proposed framework addresses the limitations of suboptimal trajectories prone to collisions, low-efficiency driving, and limited scenario validation in existing approaches. Consequently, it enhances trajectory optimization in complex and dynamic driving environments. The main contributions of this study are outlined as follows:

- A novel spatial-temporal path planning framework is proposed that integrates a DHF-based cost function into a QP-MPC structure. The cost function jointly considers safety, efficiency, comfort, and timing, addressing limitations of prior works that focus on only partial objectives [26], [27].
- The proposed cost function is embedded into the QP-MPC optimization process, enabling balanced decision-making across safety, efficiency, and comfort. Comparative experiments show it outperforms existing optimization methods in lane-changing time, control smoothness, and collision avoidance [26], [27].
- The framework is validated across diverse driving scenarios, including lane changes, overtaking, and intersection crossing. This broad evaluation goes beyond the scenario-specific validations typical in prior work and confirms the robustness and generalizability of our approach [22], [24], [26]–[28].

The rest of this study is organized as follows: Section II describes the whole framework; Section III describes DHF-based spatial trajectory planning; Section IV describes ST graph-based spatial trajectory planning; Section V describes QP-based Model Predictive control; Section VI describes

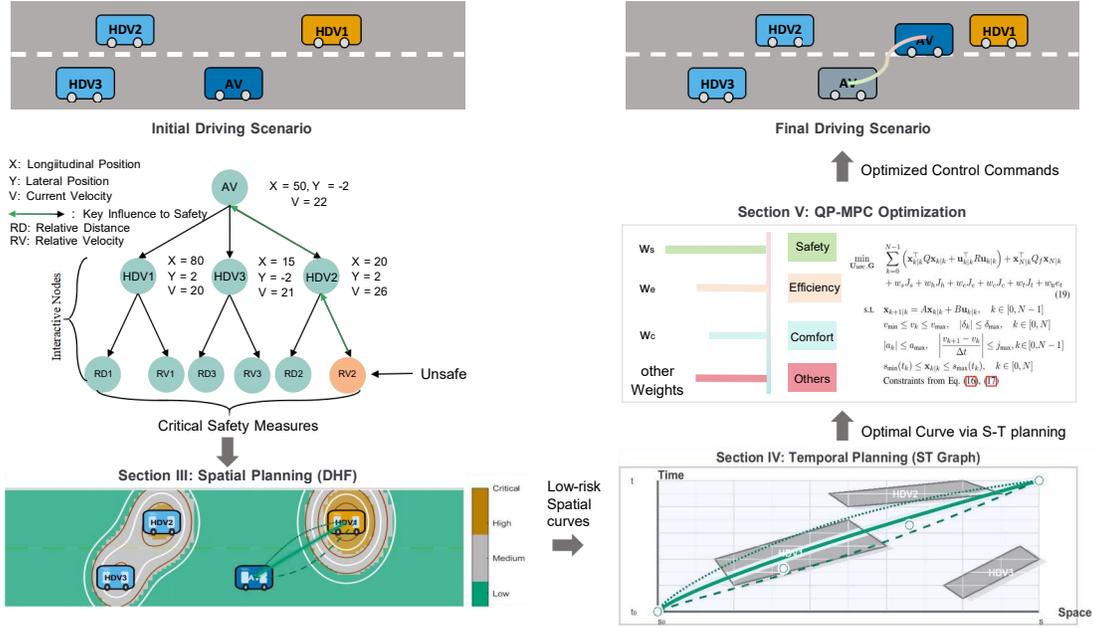

Fig. 2: The workflow of proposed framework, with the main components described in Sections III, IV, and V.

simulation results; Section VII describes QP-based Model Predictive control.

## II. System Structure

In order to solve the complexity of the trajectory planning optimization problem, this study proposes a spatial-temporal trajectory planning framework, as shown in Fig. 2.

As demonstrated in Fig. 2, four vehicles are considered in an example scenario, including one AV and three HDVs. In this scenario, the AV intends to change lanes to target lane, but there is a potential risk of collision with the yellow HDV in target lane during lane-changing. In Section III, to evaluate the risks posed in spatial domain, the HDF is used to represent the risk levels from surrounding HDVs. Additionally, a set of candidate trajectories are generated using quintic polynomials, ensuring comfort due to the smoothness of quintic polynomials.

In Section IV, the initial trajectory generated in the first stage is temporally optimized in the space-time (ST) domain using an ST graph. The future states of the surrounding HDVs are predicted and represented in the ST graph as potential collision areas, illustrated by the grey areas. The planning results must avoid overlapping with these grey areas. By using the ST graph, the spatially optimized trajectory is rechecked to ensure safety in the spatial domain.

In Section V, by integrating QP and MPC, the optimized trajectories are ensured to be spatially feasible and temporally optimized. The QP-MPC-enhanced trajectory is capable to avoid dynamic obstacles with safety, efficiency, and comforts.

After the spatial safe and comfortable planning, temporal safe planning, and safe, efficiency, and comforts-orientated optimization, the optimized trajectory is obtained.

## III. Dynamic Hazard Field-Based Spatial Trajectory Planning

This section presents an advanced path planning method that integrates a dynamic Frenet-based framework with hazard assessment and trajectory optimization. This strategy employs DHF to identify and avoid obstacles, utilizes quintic polynomials for trajectory generation, and adapts to evolving environments through a rolling horizon approach. This strategy can achieve real-time path planning in scenarios with both static and moving obstacles, ensuring compliance with vehicle kinematic constraints and passenger comfort.

To achieve this, we first establish a reference trajectory $\Xi$ as a foundational guide for optimization and hazard evaluation. By representing the vehicle's motion within a local coordinate system relative to $\Xi$, the planning process focuses on longitudinal and lateral deviations. The reference trajectory $\Xi$ is defined using a sequence of discrete points $\{(x_k, y_k)\}_{k=1}^{M}$ in the global Cartesian coordinate system, interpolated with a cubic spline to create a smooth, twice-differentiable path $p(\sigma) = [x_p(\sigma), y_p(\sigma)]^T$, where $\sigma$ is the arc length along the path. The curvature $\kappa(\sigma)$ of $\Xi$ is calculated as:

$$\kappa(\sigma) = \frac{\dot{x}_p(\sigma)\ddot{y}_p(\sigma) - \dot{y}_p(\sigma)\ddot{x}_p(\sigma)}{(\dot{x}_p(\sigma)^2 + \dot{y}_p(\sigma)^2)^{3/2}}, \quad (1)$$

where $x_p(\sigma)$ and $y_p(\sigma)$ are the coordinates of $\Xi$ as functions of $\sigma$. $\dot{x}_p(\sigma)$, $\ddot{x}_p(\sigma)$, $\dot{y}_p(\sigma)$, and $\ddot{y}_p(\sigma)$ represent the first and second derivatives of $x_p$ and $y_p$ with respect to $\sigma$.

The transformation from Frenet coordinates $(l, d)$ to Cartesian coordinates $(x, y)$ is defined as:

$$\begin{bmatrix} x \\ y \end{bmatrix} = \begin{bmatrix} x_p(l) \\ y_p(l) \end{bmatrix} + d \begin{bmatrix} -\sin\phi(l) \\ \cos\phi(l) \end{bmatrix}, \quad (2)$$

where $\phi(l) = \arctan2(\dot{y}_p(l), \dot{x}_p(l))$ is the heading angle of $\Xi$ at position $l$.

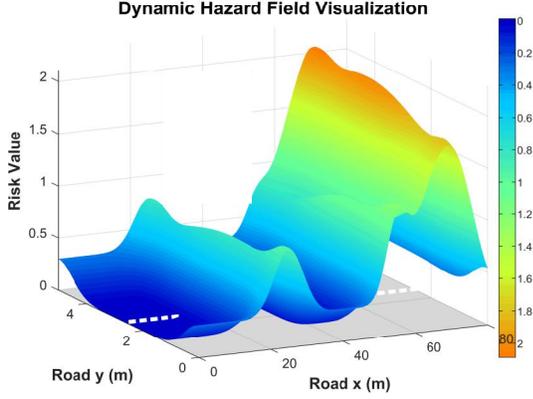

Fig. 3: Visualization of the Dynamic Hazard Field.

To ensure safe navigation in dynamic environments, we introduce the DHF, visualized in Fig. 3. The DHF quantifies the risk associated with the vehicle's current and projected positions, considering both stationary and moving obstacles.

The comprehensive hazard field $H(l, d, \tau)$ includes static hazard $H_s(l, d)$ and dynamic hazard $H_d(l, d, \tau)$:

$$H(l, d, \tau) = H_s(l, d) + H_d(l, d, \tau), \quad (3)$$

where $\tau$ represents the time duration.

The static hazard $H_s(l, d)$ accounts for road boundaries and immobile obstacles:

$$H_s(l, d) = \sum_{m=1}^{M_s} C_m \exp\left(-\frac{(l-l_m)^2}{2\eta_{l,m}^2} - \frac{(d-d_m)^2}{2\eta_{d,m}^2}\right), \quad (4)$$

where $(l_m, d_m)$ is the position of the $m$-th static obstacle or boundary point, $C_m$ is the peak hazard level for each obstacle, and $\eta_{l,m}$ and $\eta_{d,m}$ are scaling parameters defining the longitudinal and lateral hazard spread.

The dynamic hazard $H_d(l, d, \tau)$ considers the predicted trajectories of moving obstacles:

$$H_d(l, d, \tau) = \sum_{n=1}^{N_d} \frac{D_n \exp\left(-\frac{(l-l_n(\tau))^2 + (d-d_n(\tau))^2}{2\zeta_n^2(u_{rel,n})}\right)}{1 + \exp(-\lambda_n(u_{rel,n})(\Delta l_n(\tau) - \beta L_n))},$$

where $(l_n(\tau), d_n(\tau))$ denotes the predicted position of the $n$-th dynamic obstacle at time $\tau$, $D_n$ is the maximum hazard level, $u_{rel,n}$ is the relative velocity, $\zeta_n(u_{rel,n})$ adjusts the scaling factor based on $u_{rel,n}$, $\lambda_n(u_{rel,n})$ is a sensitivity parameter, $\Delta l_n(\tau)$ is the longitudinal separation from the obstacle, $L_n$ is the obstacle's length, and $\beta$ is a safety coefficient.

Fig. 4 presents visualization integrates both dynamic and static hazard fields with corresponding driving scenes in a cohesive display addressing two key scenarios: (a) highway lane-changing and (b) intersection-crossing. The top row illustrates the combined hazard field using warmer colors to denote regions of elevated risk resulting from the merging of dynamic elements, such as moving vehicles and static features like road boundaries and lane markers, while the bottom row offers a schematic top-down view of the driving scene which contextualizes vehicle positioning alongside environmental elements.

## IV. SPACE-TIME GRAPH-BASED SPATIAL TRAJECTORY PLANNING

The ST graph is a two-dimensional representation where the horizontal axis corresponds to time t, and the vertical axis denotes the longitudinal position along the reference path s. Obstacles are depicted as shaded regions based on their predicted trajectories. Let $O = \{O_1, O_2, ..., O_M\}$ represent the set of M obstacles with three driving styles: normal, aggressive, and uncertain. Each driving style is predicted using a simple assumed piecewise function:

$$s_i(t) = \begin{cases} s_i(0) + v_i(0)t & \text{Normal} \\ s_i(0) + v_i(0)t + \frac{1}{2}a_i(0)t^2 & \text{Aggressive} \\ f_{\text{random}}(s_i(0), v_i(0), a_i(0), t) & \text{Uncertain} \end{cases} \quad (5)$$

where $s_i(t)$ is the longitudinal position of obstacle $O_i$ over time t, $v_i(0)$ and $a_i(0)$ represent its initial velocity and acceleration, and $f_{\text{random}}$ predicts future positions based on the initial state.

To accommodate uncertainty in obstacle movement, a time-dependent buffer $\epsilon(t)$ is incorporated:

$$s_{\min,i}(t) = s_i(t) - \epsilon(t), \quad s_{\max,i}(t) = s_i(t) + \epsilon(t) \quad (6)$$

Each obstacle's ST region is defined as:

$$\Phi_i(t) = \{(s, t) \mid s_{\min,i}(t) \leq s \leq s_{\max,i}(t), t \in [0, T_{\max}]\} \quad (7)$$

where $s_{\min,i}(t)$ and $s_{\max,i}(t)$ denote the minimum and maximum longitudinal positions of $O_i$ at time t, respectively. The aggregate obstacle region in the ST domain is defined as:

$$\Phi_{\text{obs}} = \sum_{i=1}^{M} \Phi_i(t) \quad (8)$$

Within the ST graph, safe regions are derived as areas free from obstacles, providing safe corridors for the AV. Specifically, the safe corridor is:

$$\Phi_{\text{safe}}(t) = R \setminus \Phi_{\text{obs}}(t) \quad (9)$$

where R represents the ST domain, with each time step $t_k$ defining permissible longitudinal positions:

$$[s_{\min}(t_k), s_{\max}(t_k)] = \Phi_{\text{safe}}(t_k) \quad (10)$$

An example of the ST graph is depicted in Fig. 5. It shows a series of candidate reference speed profiles. The grey areas represent the driving positions of the HDVs, indicating that there are collision risks. However, most of the candidate profiles overlap with potential collision areas along the time axis in the ST domain. Therefore, the ST graph assists the AV in selecting a safe speed profile that avoids potential collisions.

After constructing the hazard field and ST graph, we generate a set of candidate trajectories that satisfy the vehicle's dynamic constraints, safty constrain, and adhere to the risk assessments from the DHF. These candidate trajectories are evaluated against the DHF to identify the optimal trajectory that balances safety, efficiency, and comfort.

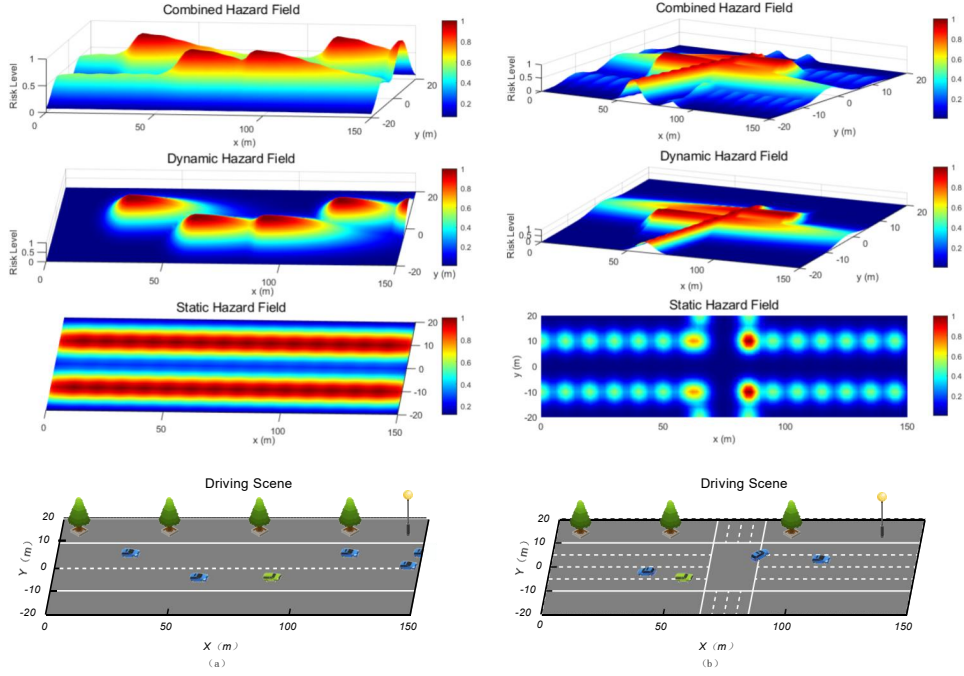

Fig. 4: Hazard field visualization: (a) highway lane-Changing, and (b) intersection-crossing scenarios.

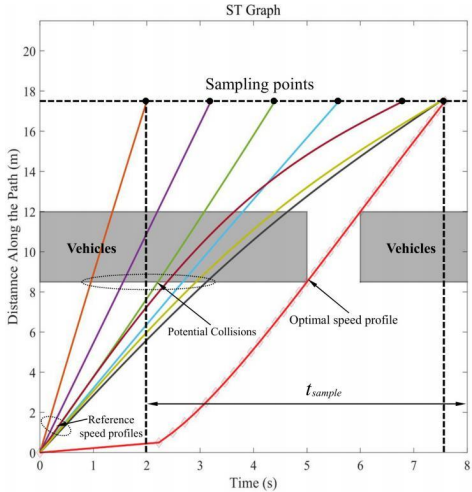

Fig. 5: Construction the ST Graph.

Each trajectory is represented using a quintic polynomial for both the longitudinal l(t) and lateral d(t) distance, ensuring smooth transitions with continuous derivatives up to the third order. The general form of the quintic polynomial is:

$$g(t) = \begin{bmatrix} 1 & 0 & 0 & 0 & 0 & 0 \\ 0 & t & 0 & 0 & 0 & 0 \\ 0 & 0 & t^2 & 0 & 0 & 0 \\ 0 & 0 & 0 & t^3 & 0 & 0 \\ 0 & 0 & 0 & 0 & t^4 & 0 \\ 0 & 0 & 0 & 0 & 0 & t^5 \end{bmatrix} \begin{bmatrix} b_0 \\ b_1 \\ b_2 \\ b_3 \\ b_4 \\ b_5 \end{bmatrix}, \quad (11)$$

where each term of the polynomial is positioned along the diagonal, where 1, t, $t^2$, $t^3$, $t^4$, and $t^5$ represent the powers of t. Each element on the diagonal is multiplied by the corresponding coefficient $b_0$, $b_1$,..., $b_5$, thus forming the polynomial g(t). The off-diagonal elements are set to zero, representing non-path regions, which visually separates the polynomial terms from inactive areas in the matrix.

To satisfy the required boundary conditions for l(t) and d(t), this paper expresses boundary conditions in matrix form as follows:

$$\mathbf{M}\,[b_0\ b_1\ b_2\ b_3\ b_4\ b_5]^T = [g_i\ g'_i\ g''_i\ g_f\ g'_f\ g''_f]^T, \quad (12)$$

where

$$\mathbf{M} = \begin{bmatrix} 1 & t_i & t_i^2 & t_i^3 & t_i^4 & t_i^5 \\ 0 & 1 & 2t_i & 3t_i^2 & 4t_i^3 & 5t_i^4 \\ 0 & 0 & 2 & 6t_i & 12t_i^2 & 20t_i^3 \\ 1 & t_f & t_f^2 & t_f^3 & t_f^4 & t_f^5 \\ 0 & 1 & 2t_f & 3t_f^2 & 4t_f^3 & 5t_f^4 \\ 0 & 0 & 2 & 6t_f & 12t_f^2 & 20t_f^3 \end{bmatrix} \quad (13)$$

where $g_i$ and $g_f$ represent the initial and final positions, $g'_i$ and $g'_f$ are the initial and final velocities, and $g''_i$ and $g''_f$ are the initial and final accelerations for both l(t) and d(t).

To generate diverse candidate trajectories, we define a sampling domain $\Phi_{sam}$ for the end states ($l_e, d_e, \tau_e$), which allows systematic variation across a feasible range of final positions, lateral deviations, and time durations:

$$\Phi = \{(l_e, d_e, \tau_e)\ |\ l_e \in [l_{min}, l_{max}],$$
$$d_e \in [d_{min}, d_{max}], \tau_e \in [\tau_{min}, \tau_{max}]\}. \quad (14)$$

For each sampled end state in $\Phi_{sam}$, we compute a corresponding trajectory by solving the polynomial coefficients in the matrix equation above. The collection of all sampled trajectories can be represented in matrix form as

**Algorithm 1** Temporal-Spatial Trajectory Planning

1: **Input:** Initial trajectory $g_0(t)$, obstacle set $O$, planning horizon $T_{max}$, vehicle and environment parameters
2: **Step 1: Spatial Risk Assessment with DHF**
3: Define DHF incorporating static and dynamic obstacles.
4: **for** each spatial position **do**
5:     Calculate DHF $H(l, d, \tau)$ in (3)
6: **end for**
7: Identify high-risk areas and feasible regions.
8: **Step 2: Temporal Planning with ST Graph**
9: Construct ST graph using obstacle predictions from $O$.
10: Define $\Phi_{safe}$ based on time-dependent obstacles in (10).
11: **Step 3: Evaluate Candidate Trajectories**
12: **for** each $g \in G$ **do**
13:     Generate path in Frenet coordinates $(l(t), d(t))$ with quintic polynomials using (11)
14:     Ensure smooth transitions with matching position, velocity, and acceleration at boundary points.
15: **end for**
16: **Output:** Final Candidate trajectory set $G$

$$G = [g_1(t) \quad g_2(t) \quad \ldots \quad g_n(t)]^T, \quad (15)$$

where $G$ is the set of all candidate trajectories generated by varying the end states $(l_e, d_e, \tau_e)$ within $\Phi_{sam}$. Each trajectory $g_k(t)$ corresponds to a unique sample in $\Phi_{sam}$ and satisfies the vehicle's boundary and dynamic constraints.

## V. QUADRATIC-PROGRAMMING-BASED MODEL PREDICTIVE CONTROL

Optimization problem for generated candidate trajectories from Section IV is defined as follows:

$$\min_G J(G) = w_s J_s + w_h J_h + w_e J_e + w_c J_c + w_t J_t + w_{tr} e_t \quad (16)$$

s.t.
$$\begin{bmatrix} l(t_i) & \dot{l}(t_i) & \ddot{l}(t_i) & l(t_f) & \dot{l}(t_f) & \ddot{l}(t_f) \\ d(t_i) & \dot{d}(t_i) & \ddot{d}(t_i) & d(t_f) & \dot{d}(t_f) & \ddot{d}(t_f) \end{bmatrix}^T = G_{boundary},$$

$$v_{min} \leq \sqrt{\dot{l}(t)^2 + \dot{d}(t)^2} \leq v_{max}, \quad |\ddot{l}(t)| \leq a_{l,max},$$

$$|\ddot{d}(t)| \leq a_{d,max}, \quad |\kappa(t)| \leq \kappa_{max},$$

$$H(l(t), d(t), t) \leq H_{max}, \quad \forall t \in [0, \tau].$$

where $J(G)$ is the total cost function integrating five sub-cost functions. The weights $w_s$, $w_h$, $w_e$, $w_c$, $w_t$, and $w_{tr}$ adjust the influence of each sub-cost function. $e_t$ is the track errors. The constraints cover speed limits, acceleration limits, curvature, and hazard exposure limits. The individual sub-cost functions are defined as:

$$J_s = \int_0^\tau (\dddot{l}(t)^2 + \dddot{d}(t)^2) dt, \quad J_h = \int_0^\tau H(l, d, t) dt,$$

$$J_e = \tau + \lambda \int_0^\tau (v_{target} - \sqrt{\dot{l}(t)^2 + \dot{d}(t)^2})^2 dt,$$

$$J_c = \int_0^\tau (\ddot{l}(t)^2 + \ddot{d}(t)^2) dt, \quad J_t = \sum_{k=1}^N t_k$$

where $J_s$ reflects smoothness through jerk minimization. $J_h$ calculates hazard exposure over time. $\dddot{d}(t)$ and $\dddot{l}(t)$ represent the snap of the lateral and longitudinal positions in Frenet coordinates, respectively, and are used to evaluate the smoothness and control quality of the trajectory. $J_e$ measures efficiency with a target speed $v_{target}$ and weighting $\lambda$. $J_c$ computes passenger comfort through minimized accelerations. $J_t$ uses total duration as a cost factor.

The vehicle's dynamic model is embedded as part of the constraints for MPC. The discrete kinematic model is:

$$x_{k+1} = x_k + v_k \cos(\theta_k) \Delta t, \quad y_{k+1} = y_k + v_k \sin(\theta_k) \Delta t,$$
$$\theta_{k+1} = \theta_k + \frac{v_k}{L} \tan(\delta_k) \Delta t, \quad v_{k+1} = v_k + a_k \Delta t, \quad (17)$$

where $(x_k, y_k)$ are the vehicle's global coordinates, $\theta_k$ is the heading angle, $v_k$ is the longitudinal velocity, $a_k$ is the acceleration, $\delta_k$ is the steering angle, $L$ is the wheelbase length, and $\Delta t$ is the discrete time step, all at step $k$.

For MPC, the vehicle model is linearized around the current state to allow for a quadratic programming (QP) approach. The linearized state-space form is:

$$x_{k+1} = A x_k + B u_k, \quad (18)$$

where $x_k = [x_k, y_k, \theta_k, v_k]^T$ is the state vector. $u_k = [a_k, \delta_k]^T$ is the control input vector.

The matrices $A$ and $B$ are derived from the linearization of the kinematic model around a nominal trajectory. The trajectory optimization is formulated as a constrained QP problem, allowing real-time control updates within MPC:

$$\min_{U_{MPC}, G} \sum_{k=0}^{N-1} \left( x_{k|k}^\top Q x_{k|k} + u_{k|k}^\top R u_{k|k} \right) + x_{N|k}^\top Q_f x_{N|k}$$
$$+ w_s J_s + w_h J_h + w_e J_e + w_c J_c + w_t J_t + w_{tr} e_t \quad (19)$$

s.t. $x_{k+1|k} = A x_{k|k} + B u_{k|k}, \quad k \in [0, N-1]$
$v_{min} \leq v_k \leq v_{max}, \quad |\delta_k| \leq \delta_{max}, \quad k \in [0, N]$
$|a_k| \leq a_{max}, \quad \left| \frac{v_{k+1} - v_k}{\Delta t} \right| \leq j_{max}, k \in [0, N-1]$
$s_{min}(t_k) \leq x_{k|k} \leq s_{max}(t_k), \quad k \in [0, N]$
Constraints from Eq. (16), (17)

where $U_{MPC} = [u_{0|k}, u_{1|k}, \ldots, u_{N-1|k}]^T$ is the sequence of control inputs over the prediction horizon. $Q$ and $R$ are weighting matrices for state and control input penalties, respectively. $Q_f$ is the terminal state penalty. $\delta_{max}$, $a_{max}$, and $j_{max}$ are maximum steering angle, acceleration, and jerk limits.

In this hierarchical optimization framework, ST optimization serves as the global planning layer, generating a reference trajectory focused on safety and smoothness. MPC acts as the local control layer, following reference trajectory in real time through short-horizon predictions. While ST sets the overall path, MPC tracks it accurately by minimizing local deviations. Importantly, ST does not depend on MPC's step size. MPC selects reference points based on its own timing, allowing both layers to function independently yet cohesively. This MPC-enhanced ST optimization ensures efficient global guidance from ST, with MPC providing robust control, thus creating an integrated and robust trajectory planning and control system.

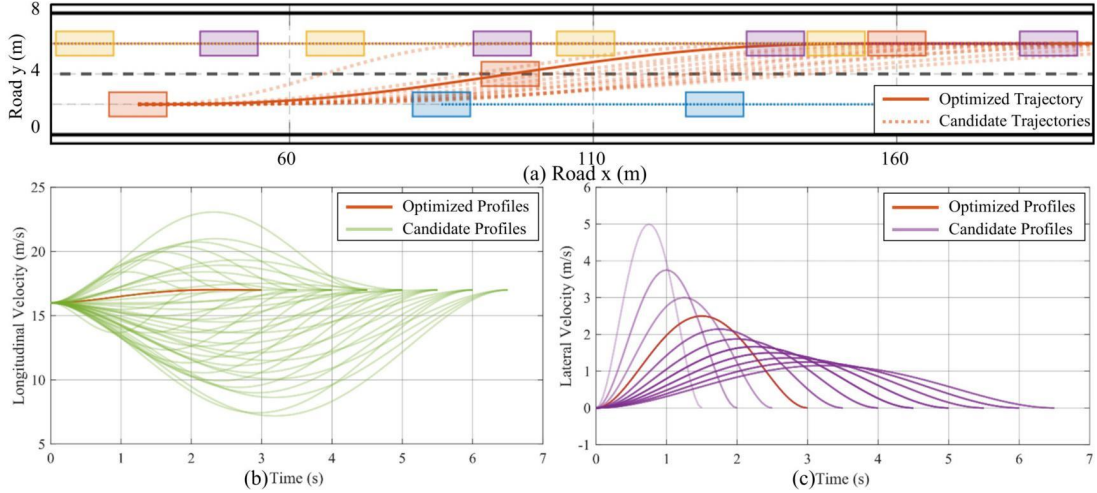

Fig. 6: The decision-making and initial planning of lane-changing scenario.

The MPC problem is structured in the QP format:

$$\min_{\Theta} \frac{1}{2}\Theta^T Q \Theta + c^T \Theta \quad (20)$$
$$\text{s.t.} \quad A_{ineq}\Theta \leq b_{ineq}, A_{eq}\Theta = b_{eq},$$

where $\Theta$ is the decision vector containing the control inputs across the prediction horizon:

$$\Theta = \begin{bmatrix} u_{0|k}^T & u_{1|k}^T & \cdots & u_{N-1|k}^T \end{bmatrix}^T. \quad (21)$$

Matrices $Q$ and $c$ encapsulate the cost terms, while $A_{ineq}$, $b_{ineq}$, $A_{eq}$, and $b_{eq}$ encode the constraints. $Q$ represents the quadratic terms in the objective function, incorporating penalties on state deviations and control efforts over the horizon:

$$Q = \text{diag}(Q, Q, \ldots, Q, Q_f, R, R, \ldots, R), \quad (22)$$

where $Q$, $R$, and $Q_f$ are the weighting matrices for state cost, control cost, and terminal state cost. $c$ encodes the linear terms in the objective function, typically zero for pure quadratic cost functions. $(A_{eq}, b_{eq})$ represents the system dynamics over the prediction horizon:

$$A_{eq} = \begin{bmatrix} I & 0 & \cdots & 0 \\ -A & I & \cdots & 0 \\ 0 & -A & \cdots & 0 \\ \vdots & \vdots & \ddots & I \\ 0 & 0 & \cdots & -A \end{bmatrix}, \quad (23)$$

where $I$ represents the identity matrix, which is a square matrix with 1 on the main diagonal and 0 elsewhere. The pair $(A_{ineq}, b_{ineq})$ defines constraints on velocity, acceleration, jerk, steering angle, and collision avoidance. These constraints are expressed as:

$$A_{ineq}\Theta \leq b_{ineq}, \quad (24)$$

where each row of $A_{ineq}$ corresponds to a specific constraint over the prediction horizon, and $b_{ineq}$ specifies the corresponding constraint bounds.

## VI. SIMULATION RESULTS

To demonstrate the effectiveness in safety and efficiency of our approach, we conduct the experiments to compare it with several advanced optimization algorithms in multiple scenarios using MATLAB 2023b. These algorithms include the proposed spatial-temporal-based QP-MPC (SQP), Differential Evolution (DE) [29], Particle Swarm Optimization (PSO) [30], the Interior Point (I_P) method [31], Genetic Algorithm (GA) [32], Pattern Search (P_S) [33], and the Active Set (A_S) method [34]. These multiple scenarios involving several surrounding vehicles (SVs) are considered to simulate real-world driving conditions and demonstrate the effectiveness of the approach. The experiments include three primary cases.

**Lane-changing Scenario:** Three SVs and one AV are on a two-lane highway. The AV starts in the lower lane behind a lead vehicle, aiming to change to the upper lane.

**Overtaking Scenario:** Two SVs and one AV are on a two-lane highway. The AV starts in the lower lane behind another vehicle, with an oncoming SV in the upper lane. The AV must execute two collision-avoided lane changes.

**Intersection-crossing Scenario:** At a four-port intersection with 2.5 m lane width, three SVs and one AV are involved. The AV starts in the lower port and must avoid SVs from different directions when changing lanes at the interSection

### A. Simulation of Lane-changing Scenario

Fig. 6 illustrates the decision-making and spatial planning of lane-changing scenario. The AV's task is to change lanes from the lower to the upper lane while interacting with the three SVs, all of which are moving from left to right. The upper subfigure illustrates the road layout with candidate and optimized trajectories, showing blue, yellow, and purple rectangles representing three SVs, respectively. The red rectangles at the bottom represent the positions of the AV. Dotted lines show various candidate trajectories for the AV, while a solid orange line with circular markers represents the optimized trajectory by the spatial planning for the AV to change lanes. The lower left and lower right subfigures display kinematic graphs, showing the longitudinal and lateral velocity during the lane change. Both the longitudinal and lateral velocity profiles are close to the median performance. This suggests that the optimized trajectory balances smoothness and efficiency compared to other candidate trajectories.

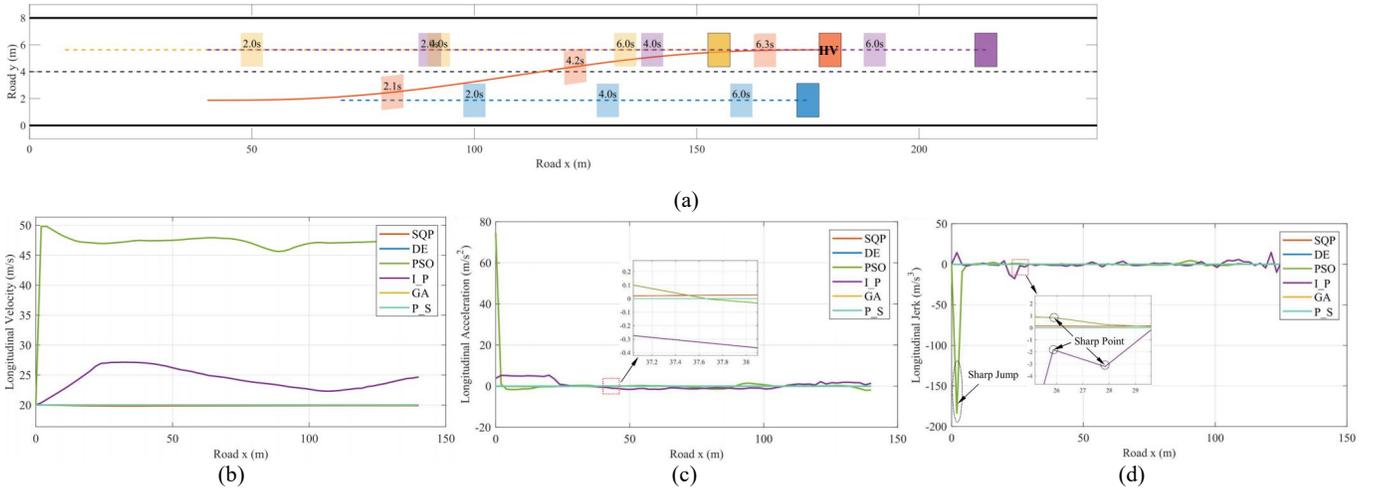

Fig. 7: The detailed planning using spatiotemporal trajectory optimization for lane-changing scenario. (a) Final optimal trajectory based on spatiotemporal planning. (b) Longitudinal velocity profiles; (c) Longitudinal acceleration profiles; (d) Longitudinal jerk profiles, all compared with advanced methods.

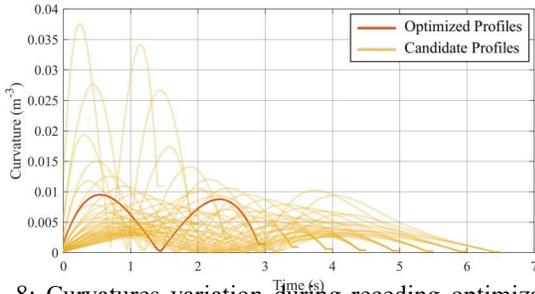

Fig. 8: Curvatures variation during receding optimization.

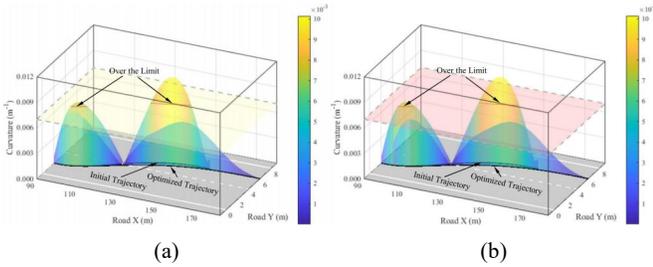

Fig. 9: Comparison of curvature variations before and after spatial and spatial-temporal optimization. (a) Results after the spatial optimization, showing the initial and optimized trajectories. The yellow plane indicates the curvature limit (b) Results after spatial-temporal optimization.

Fig. 7 illustrates the final trajectory based on the QP-MPC for lane-changing scenario, with smoothness and no collision. The x-axis represents the longitudinal distance from 0 to 200 m, while the y-axis shows the lateral position from 0 to 8 m. The trajectory of AV is depicted in red, moving from the left to the right and changing from the lower lane to the upper lane. SVs are represented by colored rectangles in both lanes. Fig. 7(b)-(d) provide a comparative analysis of the longitudinal velocity, acceleration, and jerk profiles generated by the proposed and benchmark methods: the proposed SQP, DE, PSO, I_P, GA, and P_S. For longitudinal velocity, the SQP remains a stable level, indicating a comfortable driving process. The I_P fluctuates from 23 to 27 m/s, implying reduced comfort. The PSO has a sharp lift at the initial driving, leading to a negative impact on driving experience. For longitudinal acceleration, the SQP is also more stable than these two algorithms because the I_P fluctuates around 0 and the PSO falls down sharply at the initial driving. For the longitudinal jerk, there are a series of sharp points among the I_P and PSO, while the SQP is smooth during the whole process. The DE, GA, and P_S performs well in this scenario.

Fig. 8 shows the curvature variations of candidate and optimized trajectories during a lane-changing maneuver. The red line indicates the optimized trajectory, while the yellow lines represent candidate trajectories. All trajectories exhibit two peaks, ranging from 0 to 0.04. The optimized trajectory has peaks of similar levels, indicating a smoother lane change compared to the more abrupt trajectories.

Figure 9 shows the curvature levels of the spatial and spatial-temporal trajectories in comparison to the curvature limits. Both peaks of the initial trajectory exceed the curvature limits, which negatively impacts driving experience. The trajectory considering spatial planning has curvature levels below the limits. However, the first peak is close to the limit, indicating the need for careful supervision. The spatial-temporal planning trajectory shows both peaks below the curvature limits, indicating smooth and comfortable experience.

### B. Simulation of Overtaking Scenario

Fig. 10 presents the results of spatiotemporal trajectory optimization for overtaking scenario. Fig. 10(a) shows the final trajectory of the AV in red, navigating from the left to the right and overtaking the front HDV in the lower lane. The SVs are

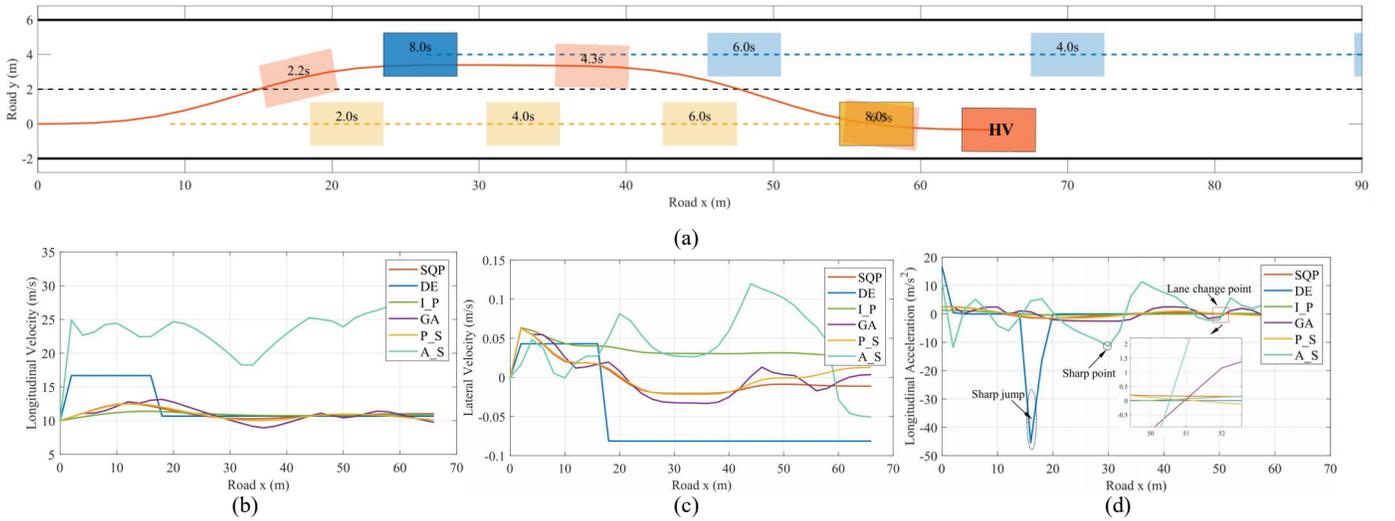

Fig. 10: The detailed planning using spatiotemporal trajectory optimization for overtaking scenario. (a) Final optimal trajectory based on spatiotemporal planning. (b) Longitudinal velocity profiles; (c) Lateral velocity profiles; (d) Longitudinal acceleration profiles, all compared with advanced methods.

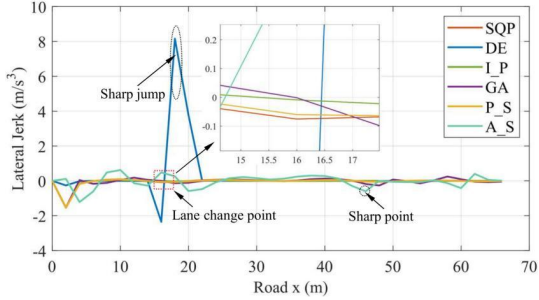

Fig. 11: Comparison of lateral jerk profiles with advanced methods in overtaking scenario.

TABLE I
COMPARISON OF ALGORITHM PERFORMANCES IN OVERTAKING SCENARIO

| Methods | $x_a$ (m/s$^2$) | $y_a$ (m/s$^2$) | $x_j$ (m/s$^3$) | $y_j$ (m/s$^3$) | T (s) |
|---|---|---|---|---|---|
| DE | 2.324 | 0.041 | 21.551 | 0.428 | **5.70 ± 0.02** |
| I_P | 0.306 | 0.014 | 0.372 | 0.057 | - |
| GA | 1.727 | 0.035 | 3.913 | 0.125 | 7.26 ± 0.03 |
| P_S | 0.920 | 0.243 | 1.096 | 0.078 | 7.27 ± 0.03 |
| A_S | 5.051 | 0.065 | 20.872 | 0.270 | 8.59 ± 0.08 |
| SQP | **0.730** | **0.0021** | **0.849** | **0.072** | 7.34 ± 0.03 |

$x_a$: average longitudinal acceleration; $y_a$: average lateral acceleration; $x_j$: average longitudinal jerk; $y_j$: average lateral jerk; T: average lane change time; -: a collision occurred.

represented by colored rectangles, with blue and yellow indicating different vehicles. The AV successfully avoids collisions with two SVs and overtakes smoothly throughout the process.

Fig. 10(b) compares longitudinal velocity profiles for the six optimization methods. The SQP exhibits the most stable velocity profile during the maneuver, ranging from 9 m/s to 13 m/s, indicating smooth driving. In contrast, the DE has a larger range, from 0 m to 18 m, suggesting instability. The A_S shows irregular changes, with peaks exceeding 25 m/s, suggesting faster but potentially less controlled driving.

Fig. 10(c) examines the lateral velocity profiles. The SQP and P_S maintain stability during the lane change, with values generally around -0.025 m/s to 0.025 m/s, indicating smooth lateral motion. The GA fluctuates between -0.03 m/s and 0.025 m/s. The I_P remains around 0.035 m/s, showing stability, while the A_S fluctuates significantly throughout the whole lean-changing, indicating instability. The DE experiences a sharp descent at 18 m, negatively affecting driving experience.

Fig. 10(d) illustrates the longitudinal acceleration. The SQP, I_P, and P_S stay close to zero throughout the process, indicating a comfortable driving. The GA shows larger fluctuations at the lane change point but remains stable during other periods. In contrast, the DE and A_S display sharp jumps and peaks, respectively, indicating reduced stability.

Fig. 11 shows the lateral jerk profiles, which are critical for evaluating the comfort of lateral movements. The SQP and P_S demonstrate a consistently smooth lateral jerk profile, with minimal variations and small peaks around -0.05 m/s$^3$, which suggests a stable and comfortable lane-change maneuver. The I_P slightly fluctuates around 0, which can also be considered as stable. The DE, GA, and A_S show sharp jumps in lateral jerk at approximately 16.5 m, 16 m, and 14.5 m, respectively. The sharp jumps indicate an abrupt movement that could result in a rough ride.

Table I compares average longitudinal acceleration ($x_a$), lateral acceleration ($y_a$), longitudinal jerk ($x_j$), lateral jerk ($y_j$), and lane change time (T) across various algorithms in overtaking scenario. The SQP outperforms others with the lowest values in multiple metrics, including a longitudinal acceleration of 0.730 m/s$^2$ and lateral acceleration of 0.0021 m/s$^2$, indicating smoother and more comfortable lane changes. In terms of jerk, the SQP also achieves a low longitudinal jerk of 0.849 m/s$^3$, which supports gradual acceleration transitions and enhances comfort. While the DE achieves the shortest lane change time at $5.70$ s, it comes with higher accelerations and jerks, suggesting a trade-off between speed and comfort. Overall, the SQP offers a balanced approach, maintaining efficient, safe, and comfortable lane changes.

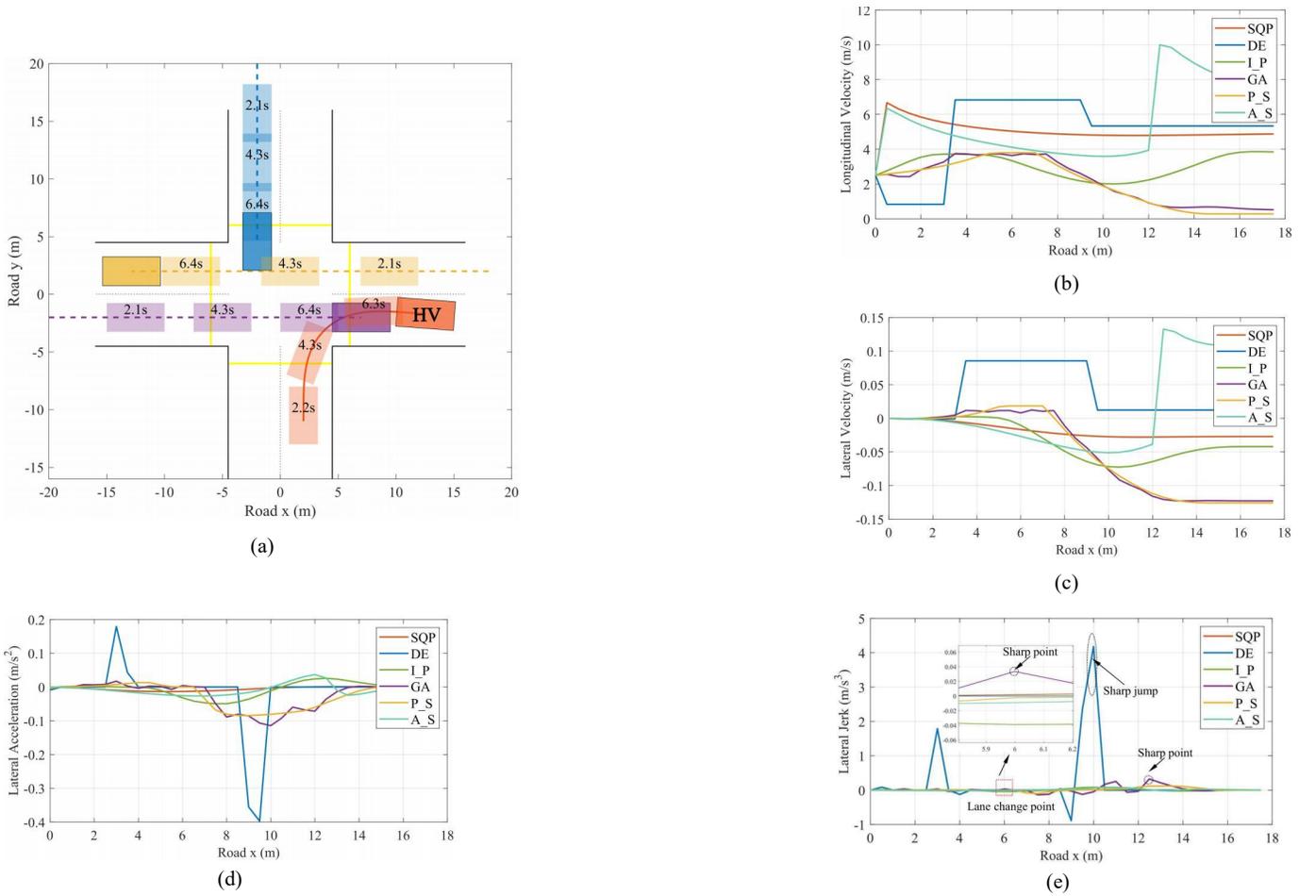

Fig. 12: The detailed planning using spatiotemporal trajectory optimization for intersection-crossing scenario. (a)Final optimal trajectory based on spatiotemporal planning. (b) Longitudinal velocity profiles; (c) Lateral velocity profiles; (d) Lateral acceleration profiles; (f) Lateral jerk profiles, all compared with advanced methods.

### C. Simulation of Intersection-crossing Scenario

Fig. 12 presents the results of spatial-temporal trajectory optimization for intersection-crossing scenario, where the AV navigates a complex intersection with multiple SVs approaching from different directions. Fig. 12(a) shows the final trajectory for the AV navigating through an intersection, interacting with three SVs, represented by colored rectangles. The yellow, blue, and purple rectangles illustrate the SVs driving from different directions. The trajectory of the AV demonstrates a smooth and safe maneuver from the bottom port to the right port. Namely, the AV successfully avoids collision with each SV, ensuring safely pass through the interSection

Fig.12 (b) compares the longitudinal velocity profiles for different optimization methods. The SQP maintains a steady speed throughout the maneuver, ranging from $5$ m/s to $7$ m/s. The DE shows large fluctuations at around $3$ m and $9$ m, indicating instability in the velocity profile. The A_S experiences a significant spike at $12$ m, which suggests a sudden increase in speed, negatively impacting comfort. The P_S and GA show gradual accelerations but has significant deviations near $8$ m, leading to reduced smooth compared to the SQP.

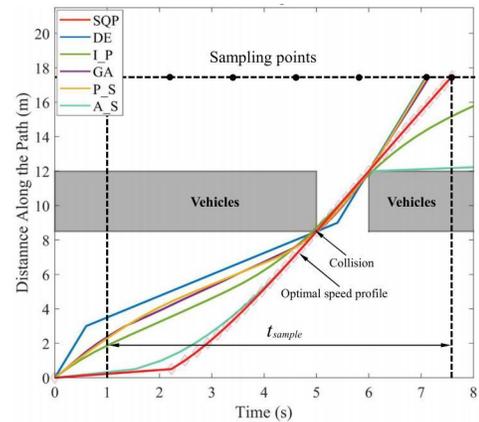

Fig. 13: Comparison of ST graph.

Fig. 12(c) shows the lateral velocity profiles. The SQP remains relatively stable with minor deviations between $-0.025$ m/s and $0$ m/s, indicating smooth lateral movement during the maneuver. On the other hand, the DE has a sharp increment and drop in lateral velocity around 3 m and 9 m, respectively, showing instability. The GA and P_S experience significant fluctuations around 7 m, which indicate reduced or less comfort. The A_S exhibits a sudden jump at 12 m, indicating an abrupt lateral maneuver.

Fig. 12(d) illustrates the lateral acceleration profiles. The SQP remains stable, fluctuating around 0 $m/s^2$, ensuring comforts and robustness. The DE has a sharp increment and drop in lateral velocity around 3 m and 9 m respectively, showing instability. The GA and P_S significantly fluctuate around the whole process. The A_S and I_P displays smooth variations in acceleration, making them less desirable for smooth driving.

Fig. 12(e) compares the lateral jerk profiles. The SQP maintains smooth jerk levels, indicating minimal sudden changes in lateral movement, which is essential for maintaining comfort during the lane change. In contrast, the DE and GA show sharp jumps and sharp points, respectively, suggesting abrupt and aggressive maneuvers, which would lower safety and comfort.

Fig. 13 depicts a comparison of speed profiles in the spatial-temporal domain. The shaded gray regions represent the positions of SVs and potential collision zones. The trajectory generated by the SQP avoids these gray areas, maintaining a safe distance from the collision regions throughout the maneuver. In contrast, the trajectory generated by the P_S, DE, and I_P cross the gray collision regions at 5 s. The trajectory generated by the A_S cross the the gray collision regions during 6 s and 7 s.

## VII. Conclusion

This study presents a novel framework for safe,efficiency,, and comfortable trajectory planning while addressesing suboptimal results, limited scenario-based verification of existing methods. By integrating DHF-enhanced QP-MPC with spatial and temporal planning, the proposed framework enhances safety, stability, and efficiency in trajectory generation. The first use of a DHF for spatial risk assessment, alongside a ST graph for temporal planning, help improve trajectory smoothness and responsiveness to dynamic obstacles. Furthermore, the introduction of a vehicle risk-assessment-based sampling method utilizing quintic polynomial curves significantly contributes to risk management and trajectory optimization. The extensive simulations demonstrate that the proposed framework ensures the safety, comfort, and smoothness of planned trajectories. Additionally, the proposed framework limits the curvatures within the curvature limits, fiurther ensuring comfortable driving. Moreover, the framework consistently outperforms popular optimization methods, showcasing superior safety, efficiency, stability, and comfort across a variety of scenarios. In the future, extensive research will be conducted in two aspects: 1) improving the predictions for HDV trajectories to enhance the accuracy of interactions, and 2) extending the algorithm to tasks involving multiple AVs by utilizing collaborative models to enhance cooperation between AVs.

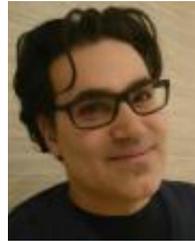

**Christos Anagnostopoulos** received the B.Sc. degree (Hons.) in computer science and telecommunications, the M.Sc. degree (Distinction) in advanced computing systems, and the Ph.D. degree in computing science from the University of Athens in 2002, 2004, and 2008, respectively. He is an Associate Professor of Distributed Computing and Data Engineering Systems and the Director of the M.Sc. Information Technology and Software Development Programmes with the School of Computing Science, University of Glasgow, Glasgow, U.K. He is leading the Knowledge and Data Engineering Systems Group (IDA Section). He is an author of over 180 scientific journals and conferences. His research expertise is at the intersection of large-scale distributed computing, distributed AI/ML, and data-centric AI.

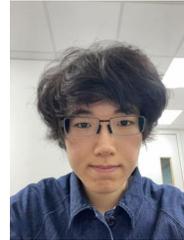

**Qiyuan Wang** received the B.Sc. degree in Computer Science and Technology from the China University of Mining and Technology, China, in 2019, and the M.Sc. degree with Distinction in Data Science from the University of Glasgow, UK, in 2020. He is currently pursuing a Ph.D. degree in Computing Science at the University of Glasgow, UK, under the supervision of Dr. Christos Anagnostopoulos. His research focuses on federated learning, edge computing, and distributed machine learning. He is currently serving as the Web & Media Chair for the 45th IEEE ICDCS 2025.

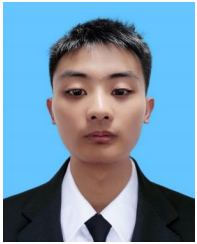

**Zhen Tian** received his bachelor degree in electronic and electrical engineering from the University of Strathclyde, Glasgow, U.K. In 2020. He is currently pursuing the Ph.D. degree with the College of Science and Engineering, University of Glasgow, Glasgow, U.K. His main research interests include Interactive vehicle decision system and autonomous racing decision systems.

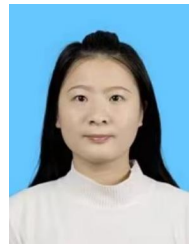

**Wenjing Zhao** received the Ph.D. degree in Traffic Engineering with Central South University, Changsha, China, in 2022. She is currently a Postdoctoral Fellow with the Department of Civil and Environmental Engineering, The Hong Kong Polytechnic University, Hong Kong, China. Her research interests include traffic safety, driving behaviour analysis, and connected vehicles.

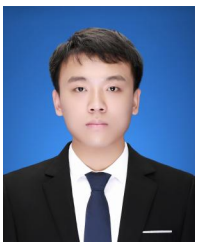

**Zhihao Lin** received the M.S. degree from the College of Electronic Science and Engineering, Jilin University, Changchun, China. He is currently pursuing the Ph.D. degree with the James Watt School of Engineering, University of Glasgow, UK. His research interests focus on multi-sensor fusion SLAM systems and robot perception in complex scenarios.

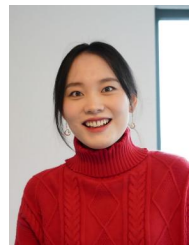

**Xiaodan Wang** received her bachelor degree in Computer Science and Technology from Xi'an Jiaotong-Liverpool University, Suzhou, China, in 2020. She is currently pursuing the Ph.D. degree with the School of Engineering, Cardiff University, Cardiff, UK. Her research interests focus on human-robot collaboration, task planning and human factors.

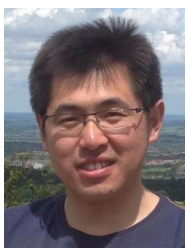

**Dezong Zhao** received the B.Eng. and M.S. degrees from Shandong University, Jinan, China, in 2003 and 2006, respectively, and the Ph.D. degree from Tsinghua University, Beijing, China, in 2010, all in Control Science and Engineering. He is a Reader in Autonomous Systems with the James Watt School of Engineering, University of Glasgow and a Turing Fellow with the Alan Turing Institute. He was awarded a Royal Society-Newton Advanced Fellow in 2020 and an EPSRC Innovation Fellow in 2018.

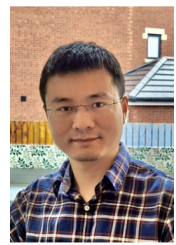

**Chongfeng Wei** received his Ph.D. degree in mechanical engineering from the University of Birmingham in 2015. He is now a Senior Lecturer (Associate Professor) at University of Glasgow, UK. His current research interests include decision-making and control of intelligent vehicles, human-centric autonomous driving, cooperative automation, and dynamics and control of mechanical systems. He is also serving as an Associate Editor of IEEE TITS, IEEE TIV, IEEE TVT, and Frontier on Robotics and AI.